\pdfoutput=1
\documentclass{article}
\usepackage{pifont}
\usepackage{comment}
\usepackage{graphicx} 
\usepackage{threeparttable}
\usepackage{amsmath}              
\DeclareMathOperator{\Concat}{Concat} 

 \usepackage[dblblindworkshop, final]{neurips_2025}
\workshoptitle{7th International Workshop on Large Scale Holistic Video Understanding: Toward Video Foundation Models}

\usepackage[utf8]{inputenc} 
\usepackage[T1]{fontenc}    
\usepackage{hyperref}       
\usepackage{url}            
\usepackage{booktabs}       
\usepackage{amsfonts}       
\usepackage{nicefrac}       
\usepackage{microtype}      
\usepackage{xcolor}         

\title{AdCare-VLM: Towards a Unified and Pre-aligned Latent Representation for Healthcare Video Understanding}

  \author{%
Md Asaduzzaman Jabin$^{1}$ \quad Hanqi Jiang$^{1}$ \quad Yiwei Li$^{1}$ \quad Patrick  Kaggwa$^{1}$ \quad Eugene Douglass$^{1}$ \\
\quad Juliet N. Sekandi$^{1}$ \quad Tianming Liu$^{1*}$ \\
$^1$University of Georgia, Athens, GA-30602, USA \\
\texttt{\{mj71006, hj67104, yl80817, pek80526, ed29157, jsekandi, tliu\}@uga.edu}\\
}

\begin{document}

\maketitle

\begin{abstract}
Chronic diseases, including diabetes, hypertension, asthma, HIV/AIDS, epilepsy, and tuberculosis, necessitate rigorous 
adherence to medication to avert disease progression, manage symptoms, and decrease mortality rates. Adherence is frequently 
undermined by factors including patient behavior, caregiver support, elevated medical costs, and insufficient healthcare 
infrastructure. We propose AdCare-VLM, a specialized LLaVA-based multimodal large vision language model (LVLM) 
by introducing a unified visual latent space with pre-alignment to facilitate visual question answering (VQA) concerning medication adherence through patient videos. We employ a private 
dataset comprising 806 custom-annotated tuberculosis (TB) medication monitoring videos, which have been labeled by clinical 
experts, to fine-tune the model for adherence pattern detection. We present LLM-TB-VQA, a detailed medical adherence VQA 
dataset that encompasses positive, negative, and ambiguous adherence cases. Our method identifies correlations between visual 
features, such as the clear visibility of the patient’s face, medication, water intake, and the act of ingestion and their associated 
medical concepts in captions. This facilitates the integration of aligned visual-linguistic representations and improves 
multimodal interactions. Experimental results indicate that our method surpasses parameter efficient fine-tuning (PEFT) 
enabled VLM models, such as LLaVA-V1.5 and Chat-UniVi, with absolute improvements ranging from 3.1\% to 3.54\% across 
pre-trained, regular, and low-rank adaptation (LoRA) configurations. Comprehensive ablation studies and attention map 
visualizations substantiate our approach, enhancing interpretability. 
\end{abstract}

\section{Introduction}
Health-risk behaviors, including medication non
adherence, impose a considerable financial burden on 
global 
healthcare 
systems. 
Preventable 
hospitalizations, supplementary treatments, and 
complications arising from untreated conditions can 
significantly elevate healthcare expenditures, with 
non-adherence rates approaching 50\% in certain 
instances [1]. This problem results in billions of 
dollars in yearly healthcare costs and is responsible for 
50\% of preventable deaths [2]. In the United States, 
non-adherence accounts for over 10\% of 
hospitalizations and approximately 125,000 deaths 
each year, resulting in significant healthcare costs. 
Tuberculosis (TB) continues to be a significant 
global health issue, accounting for 10.6 million new 
cases and 1.7 million deaths in 2021 [4]. Failure to 
adhere to tuberculosis pharmaceutical protocols 
results in treatment failures, diminished cure rates, and 
significant financial implications. The management costs associated with drug-resistant tuberculosis are 
notably elevated, highlighting the necessity for 
rigorous adherence [5]. The World Health 
Organization's End-TB strategy seeks to eradicate 
tuberculosis and its related impact by the year 2030. 
Achieving this goal is challenging due to deficiencies 
in diagnosis, treatment, and care [6]. Between 33\% 
and 50\% of patients beginning tuberculosis therapy do 
not adhere to prescribed regimens, which heightens 
the risk of drug resistance, treatment failure, extended 
infectiousness, 
and mortality, particularly in 
individuals co-infected with tuberculosis and HIV. 
Current 
adherence 
interventions 
demonstrate 
restricted efficacy [9]. In African nations, a substantial 
patient burden and significant healthcare workforce 
deficits impede effective monitoring and support for 
tuberculosis treatment [10]. Digital medication 
adherence technologies have emerged as effective 
solutions for enhancing treatment delivery in various 
healthcare settings [11]. In 2017, the World Health 
Organization (WHO) endorsed video-assisted directly 
observed therapy (VOT) as an effective alternative to 
traditional directly observed therapy (DOT) for 
monitoring tuberculosis (TB) medication adherence. 
VOT addresses geographic limitations, facilitating 
remote supervision, especially for populations that are 
difficult to access [12]. This approach empowers 
patients by enabling them to determine the timing and 
location of their medication intake [13]. Asynchronous 
VOT necessitates comprehensive manual video 
reviews to confirm daily compliance [14]. These 
evaluations adhere to a standardized protocol [15], and 
elevated patient workloads frequently result in 
repetitive tasks, thereby heightening the risk of errors 
and clinician fatigue. Excessive workload-related 
occupational stress can undermine the quality of care 
and patient outcomes [16]. Automating repetitive 
adherence monitoring tasks is essential for enhancing 
efficiency. Generative artificial intelligence (Gen-AI) 
can enhance digital adherence platforms, facilitating 
adoption and optimizing impact [17]. 

Artificial intelligence is revolutionizing healthcare, 
especially in the realm of medical imaging, by 
optimizing workflows, enhancing efficiency [19], and 
improving the experiences of healthcare workers, 
thereby reducing burnout and facilitating more direct 
patient care [20]. Generative AI (Gen-AI) is advancing 
healthcare services through improvements in clinical decision-making, population health management, and 
resource allocation [21]. Deep convolutional neural 
networks (DCNNs) enhance medical imaging, 
videography, 
and clinical deployment [22]. 
Traditional deep-learning models are largely confined  
to classification and detection tasks, facing challenges 
in generalization, contextual understanding, multi
modal reasoning, and open-ended responses [23]. 
Large Language Models (LLMs), which are trained on 
diverse datasets, demonstrate superior generalization 
capabilities, allowing them to manage a wider array of 
tasks and contexts [24].

The AI community has observed an increase in 
large language models (LLMs) such as GPT-3.5/4 
[25], PaLM [26], and BLOOM [27], which utilize 
advanced language comprehension for processing 
natural language commands. Traditional LLMs are 
limited to textual inputs, while human interaction 
encompasses various modalities, including both visual 
and textual channels. Previous studies [28-29] have 
facilitated the ability of LLMs to analyze images by 
converting them into text-like tokens. Video 
comprehension, 
although 
effective, 
presents 
significant challenges. Recent studies [30-32] have 
advanced the integration of video and language 
through shared visual encoders; however, they do not 
incorporate projection layers. X-LLM [33] and 
Macaw-LLM [34] incorporate modality-specific 
encoders; however, they exhibit inferior performance 
relative to specialized video models such as Video
ChatGPT [31], primarily due to insufficient prior 
alignment before projection. Video-LLaVA [35, 37] 
and Chat-UniVI [36] mitigate this limitation by 
integrating visual representations, aligning visual cues 
with the language feature space, and facilitating visual 
reasoning across images and videos. 

We propose a Video-LLaVA-based multimodal 
large vision-language model (LVLM) to tackle 
medication adherence challenges through visual 
question answering (VQA) in clinical settings, 
utilizing patient videos. The fine-tuned LVLM is 
trained on 806 custom-annotated videos for 
monitoring tuberculosis (TB) patient medication, 
labeled by clinical specialists. We present LLM-TB
VQA, a detailed adherence-focused VQA dataset that 
encompasses positive, negative, and ambiguous 
adherence patterns. Our approach establishes a 
correlation between visual features, including the patient's face, medication, pills, water intake, and 
ingestion actions, and their corresponding medical 
concepts in captions. This facilitates the alignment of 
visual and verbal feature representations, thereby 
improving multimodal interactions. The significance 
points are described in Table A1 (refer to Appendix). Experimental results 
indicate that our method surpasses parameter-efficient 
fine-tuning (PEFT) enabled VLMs across pre-trained, 
regular, and low-rank adaptation (LoRA) models. 
Comprehensive ablation studies and attention map 
visualizations support our methodology, enhancing 
interpretability. This study aimed to create a digital 
system that automates tele-medication monitoring and 
behavioral assessment for medication adherence.

\section{Data Source, Population and Annotation}
A multidisciplinary team consisting of three 
computer science graduate students, two machine 
learning and computer vision specialists, and 
physician-scientists with expertise in public health and 
medication adherence collaborated on this study. The 
technical feasibility of employing LVLMs to analyze 
raw video data of tuberculosis patients self
administering medication was assessed. The 
secondary dataset comprised 806 self-recorded drug 
ingestion videos collected from a pilot VOT study 
involving 51 TB patients. This pilot study took place 
in Ghana and Uganda from July 2018 to December 
2020. The study involved adult patients aged 18 to 65 
from public clinics in Kampala, Uganda, all of whom 
had a confirmed diagnosis of tuberculosis. Daily 
medication intake recordings were collected with 
written consent to assess the effectiveness of VOT in 
monitoring adherence. 

Two proficient video annotators with a background 
in public health and medical science, initially assessed the original 
dataset to create metadatas for medication intake video  
dataset. In light of the lack of established protocols for 
TB medication adherence and monitoring videos, the 
research team, consisting of three student annotators, an experienced computer scientist, and a physician 
with expertise in medication adherence, developed a 
new annotation framework through systematic review 
and discussion. The annotation process was 
categorized into three main types: positive (actual 
medication consumption), negative (absence of 
medication intake), and ambiguous (no visible tablets, 
but unclear presence of hands or face with dark and 
blurry backgrounds), as detailed in Table A2 (in Appendix). Metadata was stored in json format, including text-based video 
explanations of medication ingestion through the 
analysis of body motion (oligocentric/poly-centric), 
activities (drinking, swallowing, holding), interactions 
(talking, listening), behavioral patterns, human-object 
interactions, backgrounds, and illumination details for 
inference. 

We implemented newly established standardized 
regulations for video labeling. To guarantee the 
quality of annotations, we selected only those videos 
for which all three annotators achieved unanimous 
agreement, thereby constituting the final dataset for 
model training and evaluation. Following annotation, 
a total of 806 videos were retained, which included 
483 (60\%) cases of positive adherence, 97 (12\%) cases 
of ambiguous adherence, and 226 (28\%) cases of non
adherence. The distribution of the dataset by sex and 
adherence class is as follows: of the 483 positive 
adherence videos from 51 patients, 326 (67.4\%) were 
contributed by 28 male subjects, while 157 (32.5\%) 
were from 23 female subjects. Of the 226 non
adherence videos analyzed from 36 patients, 148 
(65.4\%) were recorded by 19 male subjects, whereas 
78 (34.6\%) were recorded by 17 female subjects. Each patient contributed approximately 10 positive and 6 
negative videos on average. We divided the dataset 
into training (70\% = 564 videos) and validation (30\% 
= 242 videos) subsets to evaluate our LVLM 
framework and baseline models for medication 
adherence recognition, question answering, and 
behavioral analysis. To mitigate the potential bias 
introduced by data imbalance, we applied a data 
balancing technique, including synthetic 
oversampling of minority classes (Adaptive-SMOTE) 
by adjusting class weights and combining these with 
undersampling majority class. During fine-tuning, assigning higher weights and temperature (>0.5) to the 
minority classes to make the model pay more attention 
to the specific imbalanced class. The class weights and 
temperature can be set through a trial and error process 
in model metadata and .init files. We conducted an 
analysis of multiple LVLM architectures for the 
purposes of detecting, classifying, and predicting 
adherence patterns. The video annotation process was 
conducted exclusively to create the dataset for model 
training and evaluation in this study.

\section{Related Work}

\subsection{Medical vision question answering (VQA)}
Early medical visual question answering (VQA) 
approaches employed general VQA algorithms [38], 
utilizing convolutional neural network (CNN) 
architectures such as VGGNet [39], ResNet [40], and 
long short-term memory (LSTM) models [41] for 
image feature extraction. In contrast, transformer
based methods like BioBert [42] were used for 
question feature extraction. Subsequently, multimodal 
fusion techniques including MCB, SAN, BAN, and 
MFB [43] were introduced. Nevertheless, the 
restricted availability of medical VQA datasets frequently resulted in overfitting and suboptimal 
performance of these models. Various strategies have 
been explored to address sample scarcity. Nguyen et 
al. [44] utilized a denoising autoencoder to optimize 
visual encoders within a meta-learning framework, 
employing unlabeled images. Do et al. [45] introduced 
a semi-annotation model designed for the pre-training 
of visual encoders, independent of external datasets. 
Previous studies employed contrastive learning 
objectives on augmented views of unannotated medical images prior to training visual encoders [46]. 
Wang et al. [47] proposed MHKD-MVQA, a 
multimodal hierarchical knowledge distillation 
method, to enhance vision-language modeling. This 
approach utilizes a novel pre-training technique on 
ROCO [48] and employs pre-trained parameters for 
the initialization of instructor and student networks. 
Eslami et al. [50] improved cross-modal 
representations by optimizing the cosine similarity 
between medical image attributes and their 
descriptions, following the initialization of their visual 
model with CLIP [49]. Ye et al. [51] proposed a caption
aware visual question-answering method that utilizes 
medical image captions for pre-training to produce 
attention maps. However, all these methods focus 
solely on medical image-based question answering, 
rather than integrating video-based adherence 
assessment. 

\begin{figure}[h!]
  \centering
  \includegraphics[width=0.8 \textwidth]{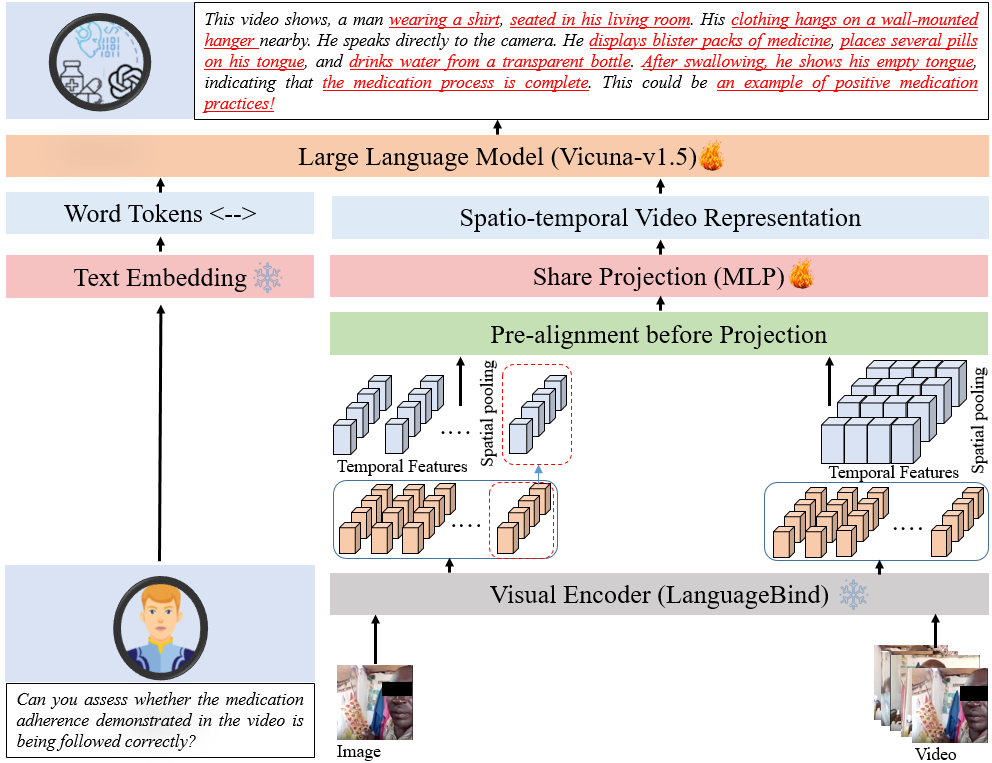} 
  \caption{Architecture: AdCare-VLM utilizes LanguageBind and a multimodal language model for video processing. The visual encoder 
extracts spatio-temporal features by averaging frame-level characteristics across both temporal and spatial dimensions. Following pre
alignment, these features are mapped into the input space of the LLM through a learnable Multi-Layer Perceptron (MLP), establishing a 
multimodal pipeline designed for effective video-based question answering in the context of medication adherence monitoring.}
  \label{fig:sample}
\end{figure}

\subsection{Large Vision Language Approach}
Expanding large language models (LLMs) to multi
modality, including photos and videos, primarily 
involves two techniques: (a) LLM as a scheduler and 
(b) LLM as a decoder. Scheduler-based methods treat 
visual models as modular components, where the 
LLM organizes tasks based on visual inputs. Our focus, however, is on the decoder-based approach 
(Table A3 in Appendix). Mini-GPT-4 [29] aligns image tokens via 
multiple layers of linear projection into an LLM input, 
but its post-alignment lacks human input. In contrast, 
mPLUG-Owl [51] employs a two-stage training: (1) 
aligning images with the language via autoregressive 
pre-training, followed by (2) instruction tuning using 
human-labeled datasets. As LLM backends scale, 
models like InstructBLIP [52] and LLaVA [35, 37, 53] 
aggregate large human instruction datasets to train 13B-parameter LVLMs, enforcing strict adherence to 
instruction-based responses and enhancing visual 
reasoning. For video comprehension, Video-ChatGPT 
[31] introduces a dataset of 100,000 video instructions, 
enabling LLMs to process video content effectively. Similarly, Video-Chat [54] and Video-LLaMA [32] 
achieve visual reasoning via collaborative training, 
allowing LLMs to handle both images and videos 
concurrently. Expanding LLMs to additional visual 
modalities often requires pre-alignment, as seen in 
LLaMA-Adapter [55] and ImageBind-LLM [56], 
where alternative modalities are integrated into the 
image space via a modality encoder. These models 
suggest that a unified feature space enhances 
multimodal reasoning. Our custom Video-LLaVA 
model differentiates itself by pre-aligning image and 
video features and performing joint training of these 
modalities. This approach enables LLMs to develop 
cohesive multimodal reasoning through a unified 
visual representation.  

\section{The Model Architecture}
\subsection{Framework Overview}
This study presents AdCare-VLM, an innovative large 
vision-language model (LVLM) that integrates 
images, videos, and text with a robust large language 
model (LLM) foundation. AdCare-VLM facilitates the 
processing of video and audio data within 
conversational contexts to monitor medication 
adherence. The model undergoes pre-training on 
VideoInstruct100K [31], which consists of 100,000 
video instructions derived from ActivityNet-200 [60], 
thereby ensuring strong performance across various 
video contexts. AdCare-VLM is developed based on 
LLaVA-1.5 [35, 37, 53] and utilizes the 
LanguageBind Model Zoo ($F_V$) [58] for the extraction 
of spatio-temporal features. It incorporates a large 
language model ($F_L$) such as Vicuna-v1.5, visual 
projection layers ($F_P$) [57], and word embeddings 
from CLIP ($F_W$) [50]. Initially, LanguageBind 
encoders convert multimodal inputs into textual 
representations, creating a cohesive visual 
representation. Subsequently, shared projection layers 
encode these features by pairing them with tokenized 
text queries, which are then input into the LLM for 
response generation. Fig. 1 presents an overview of the 
process. 

\subsection{United Visual Latent Space}
We facilitate the learning of LLMs from a unified 
vision-based representation by mapping videos and 
images into a shared feature space, positing that 
various media can transmit equivalent information. 
For instance, the act of \textit{"a person swallowing 
medication"} can be depicted using text, images, or 
videos. We compress multimodal data into a shared 
latent space to facilitate dense feature extraction and 
enhance modality interactions. We employ 
LanguageBind modality encoders [58] to align videos 
and images with the textual latent space. The architectural overview of unified representation is depicted in Fig. A1 (in Appendix).

\subsection{Pre-Alignment to Projection}
LanguageBind aligns languages and ages within a 
shared feature space automatically, using OpenCLIP 
as its initialization source [59]. The study utilizes 3 million video-text pairs from VIDAL-10M to align 
video representations with the language space [58]. 
This process facilitates the alignment of images and 
videos, integrating their representations into a unified 
visual feature space through common linguistic 
features. To connect diverse visual signals with LLM 
input, our video and image encoders are initialized 
from the LanguageBind encoder collection. A 
standard projection layer subsequently processes the 
uniform visual representation prior to its transmission 
to the LLM. 

Fig. A1 (refer to Appendix) illustrates that the image encoder and video 
encoder (LanguageBind Zoo) preprocess the selected 
frames, images, and videos to extract spatio-temporal 
features, which encompass both spatial and temporal 
characteristics. The spatial features include activity, 
texture, shape, color, object position, and place 
awareness, whereas the temporal features maintain the 
timestamp and sequence of medication activities.  The 
extracted features constitute frame-level embeddings 
that 
accurately represent the sequence of the 
medication process. The mathematical representations 
of these features are presented in Eq. 1 and 2.  We 
implemented average pooling across the spatial and 
temporal dimensions of the frame-level embeddings to 
improve the extraction of sequence-aware medication 
activity features.  The pooling operation provides a 
complete representation of the entire medication 
process sequence (see Eq. 3 and 4).  The chosen spatial 
and temporal features are subsequently aligned to 
create a cohesive visual representation (see Eq. 5). A 
shared projection layer is introduced to integrate 
multimodal capabilities.  This layer translates video
based features into the embedding space of the 
language decoder, facilitating the model's execution of 
essential video understanding tasks, including action 
recognition, event detection, and task sequence 
comprehension (see Eq. 6 and 7). The pre-aligned 
token after projection is validated through the 
optimization of likelihood probability, enabling the 
model to achieve multi-modal comprehension 
capabilities (as shown in Eq. 8). This unified visual 
representation, in conjunction with likelihood 
probability, quantifies the likelihood of the occurrence 
of specific visual features and the sequence of 
medication activities in the video. This information is 
used to identify positive, negative, and ambiguous 
medication adherence processes based on any of the visual features and task sequences (refer to Table. A2 in Appendix) 
are absent for a specific video test sample.

\subsection{Pipeline}
Overall, the LanguageBind zoo model is crucial to 
capturing the effective use of spatio-temporal 
representations in videos. And, video samples are 
represented as $V_i$,

\begin{equation}
s_i = F_V(V_i) \; \epsilon \; {\mathbb{R}^{F \times H \times W \times C}}
\end{equation}

Where, $F$ = frame count per second, $i = 1, 2, 3\ldots N;\; 
i^{th}$ frame from each second. Each $F$ frame is processed 
separately by the encoder as a series of pictures. This 
creates frame-level embedding, 

\begin{equation}
x_i \; \epsilon \; {\mathbb{R}^{F \times h \times w \times D}}
\end{equation}

Here, $h$ = height per patch = $H/p$, $w$ = width per 
patch = $W/p$, $p$ = patch size, and $N$ = $h \times w$ = total 
token count. We use average (avg) pooling over the 
temporal-dimension of embedding at frame-level to 
create a full video-level depiction, 

\begin{equation}
t_i \; \epsilon \; {\mathbb{R}^{N \times D}}
\end{equation}

Moreover, this temporal pooling method efficiently 
combines data from numerous frames. We establish 
spatial representation with average pooling along the 
spatial dimension for spatial information, 

\begin{equation}
z_i \; \epsilon \; {\mathbb{R}^{F \times D}}
\end{equation}

Last, the final latent space consists of video-level 
features and combines temporal and spatial 
characteristics from Eq. 3 and 4, as illustrated in the 
equation, 

\begin{equation}
v_i \; \epsilon \; {\mathbb{R}^{(F + N) \times D}} = [t_i, z_i]
\end{equation}

Moreover, AdCare-VLM incorporates spatio
temporal feature extraction inspired by Video-LLaVA 
[35], utilizing a higher video resolution of 224×224 
pixels to enhance frame-level encoding. We propose 
an MLP projection layer ($F_P$) to improve multimodal 
integration by mapping video-based features into the 
embedding space of the language decoder. This 
approach builds upon LLaVA-1.5 [35, 37, 53] and 
enhances multimodal capabilities beyond what is 
possible with a basic linear projection. The procedure 
produces language embedding tokens ($Q_v$) and text 
embedding ($Q_t$) computed as follows from Eq. 1-5,

\begin{equation}
Q_v = F_P(V_i) \; \epsilon \; {\mathbb{R}^{(F + N) \times K}}
\end{equation}
And,
\begin{equation}
Q_t = F_W(Q_w) \; \epsilon \; {\mathbb{R}^{L \times K}}
\end{equation}

Where, $L$ = length of the query, which is tokenized 
to ensure dimensional compatibility with video 
embeddings. The amalgamation of $Q_v$ and $Q_t$ is 
subsequently input into the vision and text decoder 
respectively, enabling the combined textual and video 
data inside the model (refer to Fig. 1). The aligned 
token before projection can be confirmed by 
optimizing the likelihood probability $p$ in Eq. 8, the 
model ultimately attains multi-modal comprehension 
skills, 

\begin{equation}
p(Q_x \mid Q_v, Q_t) = \prod_{i=1}^{L} p_{\theta}\!\left(Q_x^{[i]} \mid Q_v, Q_t^{[1:\;i-1]}\right)
\in \mathbb{R}^{(F+N+L)\times K}
\end{equation}

Here, $\theta$ and $Q_x$ refers to trainable parameters and 
generated sequences, respectively. We perform 
combined training on pictures and videos dynamically, 
with each batch comprising both text, image, and 
video data concurrently.

\subsection{Understand Pre-Training}
At this step, the model must have the capability to 
extract visual features within a comprehensive dataset 
of image and video-text pairs. Each visual feature 
correlates to an individual cycle of conversational data 
($Q_t$, $Q_a$) and $Q_a$ represents the actual reality. The 
training objective is to optimize the auto-regressive 
loss, wherein the model acquires the essential 
capability to interpret visual information. We 
immobilize the remaining model parameters during 
this step.

\subsection{Fine-tuning}
The model must deliver replies aligned with various 
instructions. These instructions frequently entail more 
intricate visual comprehension challenges rather than 
merely articulating visual messages. Be aware that the 
chat data ($Q_t^1, Q_a^1, Q_t^2, Q_a^2 \ldots Q_t^N, Q_a^N$) comprises 
several rounds, 

\begin{equation}
Q_T^m =
\begin{cases}
Q_t^1, & m = 1, \\[4pt]
\Concat\!\left(Q_t^{m-1}, Q_a^{m-1}, \ldots, Q_a^m\right), & m > 1
\end{cases}
\end{equation}

Where, $m$ = number of rounds for fine-tuning and 
when $m > 1$, we amalgamate the dialogues from all 
preceding cycles with the present directive as this 
round's input. And, training deliverable continues to be 
consistent with the prior stage. After this phase, the 
model acquires the ability to produce relevant 
responses following various directives and inquiries. 
The LLMs are also engaged in training during this 
phase.

\begin{table}[h!]
\centering
\caption{The evaluation of separated vs. joint training and alignment before the projection in AdCare-VLM (7B) with LoRA}
\begin{tabular}{ccccccc}
\hline
\textbf{Methods} & 
\multicolumn{2}{c}{\textbf{ActivityNet-200 [60]}} & 
\multicolumn{2}{c}{\textbf{VideoInstruct100K [31]}} & 
\multicolumn{2}{c}{\textbf{LLM-TB-VQA (Ours)}} \\
 & Accuracy & Score & Accuracy & Score & Accuracy & Score \\
\hline
Separated & 44.9 & 3.32 & 40.9 & 2.6 & 54.3 & 3.21 \\
Unified and aligned & 46.1 & 3.35 & 43.7 & 3.1 & 57.9 & 3.54 \\
\color{green} $\Delta$ & \color{green} +1.2 & \color{green} +0.03 & \color{green} +2.8 & \color{green} +0.5 & \color{green} +3.6 & \color{green} +0.33 \\
\hline
\end{tabular}
\label{tab:benchmark_comparison}
\end{table}

\section{Experiments and Results}
\subsection{System Details}
Our enhanced baseline is constructed using LLaVA
1.5 [35, 37, 53], incorporating LanguageBind-zoo [58] 
and CLIP [50, 59] for image encoding, alongside 
Vicuna 1.5 [57] as the language model. In the training 
process, we exclusively fine-tune the Vicuna-v1.5 and 
MLP projection layers, maintaining the integrity of the 
remaining architecture. Each video sample undergoes preprocessing to achieve a resolution of 224×224, 
comprising 8 frames per video. The model underwent 
training for five epochs, utilizing an initial learning 
rate of 1e-5, a warm-up ratio of 0.03, the AdamW 
optimizer, a cosine learning rate scheduler, and a total 
batch size of 64. The fine-tuning of our 7B model on 
eight A5000 (120GB) GPUs required approximately 
eight hours. Comprehensive information is presented 
in Table A4 (in the Appendix). The Vicuna-7B-v1.5 model serves purposes such as 
video-based conversational benchmarking, zero-shot question-answering, and the extraction of key nouns or reference expressions for the evaluation of spatial 
grounding. Furthermore, it conducts entity matching 
as cited in [57].

\begin{table}[h!]
\centering
\caption{The pre-training zero-shot performance benchmarking score of AdCare-VLM with LoRA other LVLMs (7B model)}
\begin{tabular}{p{1.4 cm}ccccccp{0.9 cm}}
\hline
 \textbf{Model} & 
\multicolumn{5}{c}{\textbf{Evaluation Metrics for Benchmark}} & \textbf{Alignment} & \textbf{Ref.} \\
 & Correctness & Detailing & Contextual & Temporal & Consistency & & \\
\hline
Video-Chat & 2.48 & 2.81 & 2.90 & 2.25 & 3.09 & \ding{55} & [54] \\
Video-LLaMA & 2.27 & 2.60 & 2.65 & 2.15 & 2.85 & \ding{55} & [32] \\
Video-ChatGPT & 2.50 & 2.51 & 2.83 & 2.32 & 3.08 & \ding{55}, $\sum$ & [31] \\
Chat-UniVi & 2.72 & 2.87 & 3.11 & 2.42 & 3.38 & \ding{55}, $<a_n>$ & [36] \\
\color{green} AdCare-VLM (Ours) & \color{green} 2.80 & \color{green} 2.93 & \color{green} 3.23 & \color{green} 2.51 & \color{green} 3.48 & \color{green}\ding{51}, $<a_n>$, $\sum$ & \color{green}- \\

\multicolumn{8}{l}{\footnotesize \textit{Note:} \ding{51} = Pre-alignment, \ding{55} = No pre-alignment, $\sum$ = Combined alignment, $<a_n>$ = Sequential alignment.}\\
\hline

\end{tabular}

\label{tab:videochat_eval}
\end{table}

\subsection{Evaluation Process}
We evaluate LLM-TB-VQA and other video 
datasets to assess the impact of prior alignment before 
projection and joint training, utilizing both separated 
and unified training approaches. In the separated 
approach, MAE [61] functions as the visual representation, whereas in the unified approach, the 
LanguageBind encoder delivers a combined visual 
representation. We replace the vision encoder with one 
of the same scale, while maintaining the 
LanguageBind video encoder for integrated training. 
As shown in Table 1, the unified visual representation 
significantly improves performance across four VQA 
datasets. In contrast, isolated visual representations 
demonstrate reduced accuracy and display comparable 
accuracy patterns. The findings indicate that the 
integration of visual representations improves the capacity of LVLMs to learn from and understand video data.

This section provides a summary of the quantitative 
assessments that evaluate the impact of the enhanced 
baseline on AdCare-VLM. The benchmarking 
framework from Video-ChatGPT [31] is utilized to 
assess performance across critical dimensions: 
accuracy, attention to detail, contextual 
comprehension, temporal awareness, and consistency. 
To enhance reproducibility, we substitute GPT-4.0 with Vicuna-7B-v1.5 [57], thereby addressing 
the repeatability limitations associated with proprietary models. Subsequently, we reassess 
AdCare-VLM alongside contemporary models to 
ensure a fair comparison. Table 2 indicates that 
AdCare-VLM exceeds the performance of Video
ChatGPT and Chat-UniVi, addressing recent 
advancements while effectively tackling token 
alignment and fine-tuning issues. In the assessment of 
LLM-TB-VQA fine-tuning, AdCare-VLM is 
compared solely with Chat-UniVi, its main 
competitor, owing to limited pre-alignment issues and enhanced benchmark performance (refer to Table 2). 
The results demonstrate AdCare-VLM's superiority, 
especially in contextual and temporal understanding, 
compared to both foundational and advanced models. 
Table A5 (in Appendix) presents the performance of AdCare-VLM 
across pre-trained, regular, and LoRA fine-tuning with 
Chat-UniVi. Qualitative results depicted in Fig. A2 (Appendix) 
demonstrate 
the 
model's 
enhanced baseline 
performance. Further demonstrations are included in 
the Appendix (Fig. A3 and A4). The AdCare-VLM-7B 
model utilizing LoRA demonstrates improved 
accuracy, greater descriptive detail, and enhanced 
contextual alignment with the temporal progression of 
medication videos. This representation more 
accurately reflects patient interactions with their 
environment and adherence behaviors, demonstrating 
a nuanced understanding of actions and contexts. The 
qualitative insights align with the quantitative results, 
emphasizing the improved baseline's significance in 
AdCare-VLM's video comprehension.

\section*{Ethical Approvals}
Institutional Review Board Office of Research at 
the University of Georgia (PROJECT00002406) and 
Makerere University Higher Degrees, Research, and 
Ethics Committee in Uganda (756) approved the 
project. 

\section*{Source Code and Data Availability}
The code and pre-trained weights will be accessible 
at: \url{https://github.com/asad14053/AdCare-VLM}.

Due to human subject’s privacy limitations, the 
datasets created and/or analyzed during the current 
study are not publicly available; however, it can be 
obtained from the corresponding author upon 
reasonable request if approved by institution’s IRB. 

\section*{Conclusion}
This research presents an innovative medical Large 
Vision Language Model (LVLM) methodology by utilizing unified and pre-alignment of latent space for LLaVA-V1.5 and Vicuna-V1.5, aimed at 
concurrently 
acquiring 
visual 
and 
textual 
representations through a self-supervised approach. 
The validation occurs in downstream medical Visual 
Question Answering (VQA) for medication adherence 
detection, tackling issues like healthcare personnel 
shortages, the necessity for assistance, and patient 
burden. The proposed system undergoes pre-training 
on two publicly accessible activity datasets and is 
subsequently fine-tuned on LLM-TB-VQA utilizing 
an 
alignment-before-projection 
methodology. 
Experimental results indicate that our method 
demonstrates superior performance compared to state
of-the-art algorithms across four LVLM models. We 
demonstrate the potential of vision-language models 
and generative AI for predicting medication adherence 
through a unique video dataset derived from the 
African context. 

\section*{References}
{
\small
[1] Lee, E. K., Poon, P., Yip, B. H., Bo, Y., Zhu, M. T., Yu, C. P., 
... \& Wong, S. Y. (2022). Global burden, regional differences, 
trends, and health consequences of medication nonadherence 
for hypertension during 2010 to 2020: a meta‐analysis 
involving 27 million patients, {\it Journal of the American Heart 
Association}, 11(17), e026582.

[2] Phillips, L. A., Pluta, K., \& More, K. R. (2022). Health
Related Behaviours–Treatment Non-Adherence. In Health
Related Behaviours: Treatment Non-Adherence: {\it Routledge 
Encyclopaedia of Psychology in the Real World}. Routledge.  

[3] Kumaraswamy, M., Samraksha, M., Meghana, M. N., \& 
Shreeharsha, P. B. (2022). A review on barriers of medication 
adherence in chronic diseases, {\it International Journal of 
Pharmaceutical Sciences Review and Research}, 75(2), 17–26. 
\url{https://doi.org/10.47583/ijpsrr.2022.v75i02.003}. 

[4] World Health Organization. (2022, October 27). {\it Global tuberculosis report 2022.} 
\url{https://www.who.int/publications/i/item/9789240061729}.

[5] Chimeh, R. A., Gafar, F., Pradipta, I. S., Akkerman, O. W., 
Hak, E., Alffenaar, J. W., \& Van Boven, J. F. M. (2020). 
Clinical and economic impact of medication non-adherence in 
drug-susceptible tuberculosis: a systematic review, {\it The international Journal of Tuberculosis and Lung Disease}, 24(8), 
811-819. 

[6] World Health Organization. (2015). Digital health for the End 
TB Strategy: an agenda for action (No. 
WHO/HTM/TB/2015.21). {\it World Health Organization.}

[7] Anuwatnonthakate, A., Limsomboon, P., Nateniyom, S., 
Wattanaamornkiat, W., Komsakorn, S., Moolphate, S., ... \& 
Varma, J. K. (2008). Directly observed therapy and improved 
tuberculosis treatment outcomes in Thailand, {\it PLoS One}, 3(8), 
e3089. 

[8] Waitt, C. J., \& Squire, S. B. (2011). A systematic review of 
risk factors for death in adults during and after tuberculosis 
treatment, {\it The International journal of tuberculosis and lung 
disease}, 15(7), 871-885. 

[9] Alipanah, N., Jarlsberg, L., Miller, C., Linh, N. N., Falzon, D., 
Jaramillo, E., \& Nahid, P. (2018). Adherence interventions 
and outcomes of tuberculosis treatment: A systematic review 
and meta-analysis of trials and observational studies, {\it PLoS 
medicine}, 15(7), e1002595. 

[10] Bulage, L., Sekandi, J., Kigenyi, O., \& Mupere, E. (2014). The 
quality of tuberculosis services in health care centres in a rural 
district in Uganda: the providers’ and clients’ perspective, 
{\it Tuberculosis research and treatment}, 2014(1), 685982.

[11] World Health Organization. (2015). New horizons for health 
through mobile technologies: second global survey on 
eHealth. Geneva, Switzerland: {\it World Health Organization}; 
2011.

[12] Garfein, R. S., \& Doshi, R. P. (2019). Synchronous and 
asynchronous video observed therapy (VOT) for tuberculosis 
treatment adherence monitoring and support, {\it Journal of 
clinical tuberculosis and other mycobacterial diseases}, 17, 
100098. 

[13] Garfein, R. S., Liu, L., Cuevas-Mota, J., Collins, K., Muñoz, 
F., Catanzaro, D. G., ... \& Raab, F. (2018). Tuberculosis 
treatment monitoring by video directly observed therapy in 5 
health districts, California, USA, {\it Emerging infectious 
diseases}, 24(10), 1806.

[14] Garfein, R. S., Liu, L., Cuevas-Mota, J., Collins, K., 
Catanzaro, D. G., Muñoz, F., ... \& Rios, P. (2020). Evaluation 
of recorded video-observed therapy for anti-tuberculosis 
treatment, {\it The International Journal of Tuberculosis and Lung 
Disease}, 24(5), 520-525. 

[15] Centers for Disease Control and Prevention. (n.d.). 
Implementing an electronic directly observed therapy (eDOT) 
program: A toolkit for tuberculosis programs. {\it U.S. 
Department of 
Health \& Human Services.} \url{https://www.cdc.gov/tb-programs/php/edot
toolkit/index.html}.

[16] Erickson, S. M., Rockwern, B., Koltov, M., McLean, R. M., \& 
Medical Practice and Quality Committee of the American 
College of Physicians (2017). Putting patients first by 
reducing administrative tasks in health care: a position paper 
of the American College of Physicians, {\it Annals of internal 
medicine}, 166(9), 659-661.

[17] Doshi, R., Falzon, D., Thomas, B. V., Temesgen, Z., 
Sadasivan, L., Migliori, G. B., \& Raviglione, M. (2017). 
Tuberculosis control, and the where and why of artificial 
intelligence, {\it ERJ open research}, 3(2).

[18] Pinto-Coelho, L. (2023). How artificial intelligence is shaping 
medical imaging technology: A survey of innovations and 
applications, {\it Bioengineering}, 10(12), 1435.

[19] Bajwa, J., Munir, U., Nori, A., \& Williams, B. (2021). 
Artificial intelligence in healthcare: transforming the practice 
of medicine, {\it Future healthcare journal}, 8(2), e188-e194. 

[20] Pavuluri, S., Sangal, R., Sather, J., \& Taylor, R. A. (2024). 
Balancing act: the complex role of artificial intelligence in 
addressing burnout and healthcare workforce dynamics, {\it BMJ 
Health \& Care Informatics}, 31(1), e101120. 

[21] Falzon, D., Migliori, G., Jaramillo, E., \& Raviglione, M. 
Digital health technology for the end TB strategy: Developing 
priority products and making them work.

[22] Sekandi, J. N., Shi, W., Zhu, R., Kaggwa, P., Mwebaze, E., \& 
Li, S. (2023). Application of artificial intelligence to the 
monitoring of medication adherence for tuberculosis treatment 
in Africa: algorithm development and validation, {\it JMIR AI}, 
2(1), e40167. 

[23] Bay, Y. Y., \& Yearick, K. A. (2024). Machine Learning vs 
Deep Learning: The Generalization Problem, {\it arXiv preprint},  
arXiv:2403.01621. 

[24] Sun, Y., \& Liu, X. (2025). Research and Application of a 
Multi-Agent-Based Intelligent Mine Gas State Decision
Making System, {\it Applied Sciences}, 15(2), 968. 

[25] Achiam, J., Adler, S., Agarwal, S., Ahmad, L., Akkaya, I., 
Aleman, F. L., ... \& McGrew, B. (2023). Gpt-4 technical 
report, {\it arXiv preprint}, arXiv:2303.08774.

[26] Bi, B., Li, C., Wu, C., Yan, M., Wang, W., Huang, S., ... \& Si, 
L. (2020). Palm: Pre-training an autoencoding \& autoregressive 
language model for context-conditioned generation, {\it arXiv 
preprint}, arXiv:2004.07159. 

[27] Le Scao, T., Fan, A., Akiki, C., Pavlick, E., Ilić, S., Hesslow, 
D., ... \& Al-Shaibani, M. S. (2023). Bloom: A 176b-parameter 
open-access multilingual language model.

[28] Ye, J., Hu, A., Xu, H., Ye, Q., Yan, M., Dan, Y., ...\& Huang, 
F. (2023). mplug-docowl: Modularized multimodal large 
language model for document understanding, {\it arXiv preprint}, arXiv:2307.02499. 

[29] Zhu, D., Chen, J., Shen, X., Li, X., \& Elhoseiny, M. (2023). 
Minigpt-4: Enhancing vision-language understanding with 
advanced large language models, {\it arXiv preprint}, arXiv:2304.10592.

[30] Alayrac, J. B., Donahue, J., Luc, P., Miech, A., Barr, I., 
Hasson, Y., ... \& Simonyan, K. (2022). Flamingo: a visual 
language model for few-shot learning. Advances in neural 
information processing systems, 35, 23716-23736. 

[31] Maaz, M., Rasheed, H., Khan, S., \& Khan, F. S. (2023). Video
chatgpt: Towards detailed video understanding via large vision 
and language models, {\it arXiv preprint}, arXiv:2306.05424.

[32] Zhang, H., Li, X., \& Bing, L. (2023). Video-llama: An 
instruction-tuned audio-visual language model for video 
understanding, {\it arXiv preprint}, arXiv:2306.02858.

[33] Chen, F., Han, M., Zhao, H., Zhang, Q., Shi, J., Xu, S., \& Xu, 
B. (2023). X-llm: Bootstrapping advanced large language 
models by treating multi-modalities as foreign languages, {\it arXiv preprint}, arXiv:2305.04160.

[34] Lyu, C., Wu, M., Wang, L., Huang, X., Liu, B., Du, Z., ... \& 
Tu, Z. (2023). Macaw-llm: Multi-modal language modeling 
with image, audio, video, and text integration, {\it arXiv preprint}, arXiv:2306.09093. 

[35] Lin, B., Ye, Y., Zhu, B., Cui, J., Ning, M., Jin, P., \& Yuan, L. 
(2023). Video-llava: Learning united visual representation by alignment before projection, {\it arXiv preprint}, arXiv:2311.10122.

[36] Jin, P., Takanobu, R., Zhang, W., Cao, X., \& Yuan, L. (2024). 
Chat-univi: Unified visual representation empowers large 
language models with image and video understanding. In 
Proceedings of the IEEE/CVF Conference on Computer 
Vision and Pattern Recognition (pp. 13700-13710).

[37] Munasinghe, S., Thushara, R., Maaz, M., Rasheed, H. A., 
Khan, S., Shah, M., \& Khan, F. (2023). Pg-video-llava: Pixel 
grounding large video-language models, {\it arXiv preprint}, arXiv:2311.13435.

[38] Abacha, A. B., Gayen, S., Lau, J. J., Rajaraman, S., \& Demner
Fushman, D. (2018, September). NLM at ImageCLEF 2018 
Visual Question Answering in the Medical Domain, {\it In CLEF}, (working notes) (pp. 1-10).

[39] Simonyan, K. (2014). Very deep convolutional networks for large-scale image recognition, {\it arXiv 
preprint}, arXiv:1409.1556. 
 
[40] He, K., Zhang, X., Ren, S., \& Sun, J. (2016). Deep residual 
learning for image recognition, {\it In Proceedings of the IEEE 
conference on computer vision and pattern recognition}, (pp. 
770-778). 

[41] Hochreiter, S. (1997). Long Short-term Memory, {\it Neural 
Computation MIT-Press}.

[42] Lee, J., Yoon, W., Kim, S., Kim, D., Kim, S., So, C. H., \& 
Kang, J. (2020). BioBERT: a pre-trained biomedical language 
representation model for biomedical text mining, {\it Bioinformatics}, 36(4), 1234-1240.

[43] Yu, Z., Yu, J., Fan, J., \& Tao, D. (2017). Multi-modal 
factorized bilinear pooling with co-attention learning for 
visual question answering, {\it In Proceedings of the IEEE 
international conference on computer vision}, (pp. 1821-1830).

[44] Nguyen, B. D., Do, T. T., Nguyen, B. X., Do, T., Tjiputra, E., 
\& Tran, Q. D. (2019). Overcoming data limitation in medical 
visual question answering, {\it In Medical Image Computing and 
Computer Assisted Intervention–MICCAI 2019: 22nd 
International Conference}, Shenzhen, China, October 13–17, 
2019, Proceedings, Part IV 22 (pp. 522-530). Springer 
International Publishing. 

[45]  Do, T., Nguyen, B. X., Tjiputra, E., Tran, M., Tran, Q. D., \& 
Nguyen, A. (2021). Multiple meta-model quantifying for 
medical visual question answering, {\it In Medical Image 
Computing and Computer Assisted Intervention–MICCAI 
2021: 24th International Conference}, Strasbourg, France, 
September 27–October 1, 2021, Proceedings, Part V 24 (pp. 
64-74). Springer International Publishing.

[46] Liu, B., Zhan, L. M., Xu, L., \& Wu, X. M. (2022). Medical 
visual question answering via conditional reasoning and 
contrastive learning, {\it IEEE transactions on medical imaging}, 
42(5), 1532-1545. 

[47]  Wang, J., Huang, S., Du, H., Qin, Y., Wang, H., \& Zhang, W. 
(2022, December). Mhkd-mvqa: Multimodal hierarchical 
knowledge distillation for medical visual question answering, {\it In 2022 IEEE International Conference on Bioinformatics and 
Biomedicine (BIBM)}, (pp. 567-574). IEEE.

[48]  Pelka, O., Koitka, S., Rückert, J., Nensa, F., \& Friedrich, C. 
M. (2018). Radiology objects in context (roco): a multimodal 
image dataset. In Intravascular Imaging and Computer 
Assisted Stenting and Large-Scale Annotation of Biomedical 
Data and Expert Label Synthesis: {\it 7th Joint International 
Workshop, CVII-STENT 2018 and Third International 
Workshop, LABELS 2018, Held in Conjunction with 
MICCAI 2018}, Granada, Spain, September 16, 2018, 
Proceedings 3 (pp. 180-189). Springer International 
Publishing. 

[49] Radford, A., Kim, J. W., Hallacy, C., Ramesh, A., Goh, G., 
Agarwal, S., ... \& Sutskever, I. (2021, July). Learning 
transferable visual models from natural language supervision, {\it In International conference on machine learning}, (pp. 8748
8763), PMLR.

[50] Eslami, S., Meinel, C., \& De Melo, G. (2023, May). 
Pubmedclip: How much does clip benefit visual question answering in the medical domain?, {\it In Findings of the 
Association for Computational Linguistics: EACL 2023}, (pp. 
1181-1193). 

[51] Ye, Q., Xu, H., Xu, G., Ye, J., Yan, M., Zhou, Y., ... \& Zhou, 
J. (2023). mplug-owl: Modularization empowers large 
language models with multimodality, {\it arXiv preprint}, arXiv:2304.14178. 

[52] Dai, W., Li, J., Li, D., Tiong, A. M. H., Zhao, J., Wang, W., ... 
\& Hoi, S. (2023). Instructblip: Towards general-purpose 
vision-language models with instruction tuning, {\it arXiv preprint}, arXiv:2305.06500. 

[53] Lin, B., Tang, Z., Ye, Y., Cui, J., Zhu, B., Jin, P., ... \& Yuan, 
L. (2024). Moe-llava: Mixture of experts for large vision
language models, {\it arXiv preprint}, arXiv:2401.15947. 

[54] Li, K., He, Y., Wang, Y., Li, Y., Wang, W., Luo, P., ... \& Qiao, 
Y. (2023). Videochat: Chat-centric video understanding, {\it arXiv preprint}, arXiv:2305.06355. 

[55] Zhang, R., Han, J., Liu, C., Gao, P., Zhou, A., Hu, X., ... \& 
Qiao, Y. (2023). Llama-adapter: Efficient fine-tuning of 
language models with zero-init attention, {\it arXiv preprint}, arXiv:2303.16199. 

[56] Han, J., Zhang, R., Shao, W., Gao, P., Xu, P., Xiao, H., ... \& 
Qiao, Y. (2023). Imagebind-llm: Multi-modality instruction 
tuning, {\it arXiv preprint}, arXiv:2309.03905.

[57] Chiang, W. L., Li, Z., Lin, Z., Sheng, Y., Wu, Z., Zhang, H., 
... \& Xing, E. P. (2023). Vicuna: An open-source chatbot 
impressing gpt-4 with 90\%* chatgpt quality. See 
\url{https://vicuna.lmsys.org}, (accessed 14 April 2023), 2(3), 6.

[58] Zhu, B., Lin, B., Ning, M., Yan, Y., Cui, J., Wang, H., ... \& 
Yuan, L. (2023). Languagebind: Extending video-language 
pretraining to n-modality by language-based semantic 
alignment, {\it arXiv preprint}, arXiv:2310.01852. 

[59] Ilharco, G., Wortsman, M., Carlini, N., Taori, R., Dave, A., 
Shankar, V., Namkoong, H., Miller, J., Hajishirzi, H., Farhadi, 
A., \& Schmidt, L. (2021). OpenCLIP (0.1). Zenodo. 
\url{https://doi.org/10.5281/zenodo.5143773}.

[60] Caba Heilbron, F., Escorcia, V., Ghanem, B., \& Carlos 
Niebles, J. (2015). Activitynet: A large-scale video benchmark 
for human activity understanding, {\it In Proceedings of the ieee 
conference on computer vision and pattern recognition}, (pp. 
961-970). 

[61] He, K., Chen, X., Xie, S., Li, Y., Dollár, P., \& Girshick, R. 
(2022). Masked autoencoders are scalable vision learners, {\it In 
Proceedings of the IEEE/CVF conference on computer vision 
and pattern recognition}, (pp. 16000-16009).


\appendix

\section{Technical Appendices and Supplementary Material}

\subsection{Limitation and Future Direction}
AdCare-VLM 
demonstrates 
competitive 
performance in both photos and videos; however, it 
still has several limitations. The necessity for open
access, large-scale annotated datasets, particularly in 
the African context, is essential for advancing 
research. Moreover, it is essential to rectify disparities in data distribution to enhance the model's efficacy.  
The SMOTE strategy for over-sampling the minority 
class may occasionally result in overfitting.  In 
addition to technological constraints, dataset biases 
such as gender, socio-economic, and cultural biases 
provide a substantial obstacle to equitable and efficient 
model implementation.  Conversely, the acquisition of 
more clean and organized data is costly, labor
intensive, and fraught with regulatory complexities.  
We want to adopt cluster-based SMOTE together with bias-aware fine-tuning and assessment to address data 
shortages and possible biases in the future. 

The 
extensive 
implementation of LVLM 
technology is contingent upon variables including data 
storage, computational resources, and healthcare 
infrastructure. 
AdCare-VLM exhibits moderate 
capability in understanding extended videos from a 
modeling standpoint. Table A5 (in Appendix) indicates that Chat
UniVi outperforms on our pre-trained AdCare-VLM 
by 0.2 on LLM-TB-VQA, largely because AdCare
VLM depends on uniformly sampled 8 frames, 
potentially missing complex details in longer videos. 
Furthermore, the pre-training of AdCare-VLM 
demands significant resources, necessitating 3–4 days 
on 8×A5000-120G GPUs. It can serve as a basis for 
extending to other visual modalities, including depth 
and infrared imagery. Future research may investigate 
the integration of timestamps, object segmentation, 
and grounding embeddings to improve the capacity of 
visual-language 
models 
to 
relationships in their responses.

\renewcommand{\thefigure}{A\arabic{figure}}
\setcounter{figure}{0} 
\renewcommand{\thetable}{A\arabic{table}}
\setcounter{table}{0} 

\begin{table}[h!]
\centering
\caption{Statement of significance}
\begin{tabular}{l p{10cm}}
\toprule
\textbf{Topic} & \textbf{Description} \\
\midrule
Research Questions & 
\begin{itemize}
 \item The absence of extensively annotated 
datasets complicates the training of VQA 
models. This lack of data makes it difficult 
to construct models that accurately analyze 
medical images and answer associated 
questions.
 \item Noisy tasks like triple extraction and 
relation extraction might impair LLM 
performance by introducing extraneous 
information in incoming phrases. 
\end{itemize}
 \\
What is already 
known  & 
\begin{itemize}
\item Contemporary medical VQA models only 
utilize picture captioning datasets for pre
training before fine-tuning for VQA tasks. 
\item LLM often suffers from hallucinations and 
causes cacophony while dealing with public 
health datasets. 
Output terminal
\end{itemize}\\
What this paper 
adds  & 
\begin{itemize}
    \item Our custom AdCare-VLM model can 
outperform simultaneous visual reasoning 
skills for both photos and movies.
    \item Our AdCare-VLM unifies visual 
representations and offers to align before 
projection by binding visual signals to 
linguistic 
feature 
space to prevent 
hallucination and cacophony.
\item This technique uses a few fine-training 
objectives- masked video token modeling, 
text token modeling, video-text token 
alignment, and video-caption contrastive 
learning to better align visual features with 
written notions. 
\item Downstreaming task for tele-medication 
and introducing huggingface eco-system.
\end{itemize} \\
\bottomrule
\end{tabular}
\end{table}

\begin{table}[h!]
\centering
\caption{ The rules for populating the LLM-TB-VQA video dataset and video annotations}
\centering
\begin{tabular}{c p{7cm} c}
\toprule
\textbf{Data Annotation} & \centering\textbf{Description} & \textbf{Distribution} (Total- 806) \\
\midrule
\textbf{Adherence} &  Videos show the face, pill, and water bottle clearly, patient takes medications and drinks water, and good lighting. & 60\% \\
\textbf{Non-adherence}  & No patient face was seen, no tablets found, no pill intake or water intake. & 28\% \\
\textbf{Ambiguous}  &  Pills not visible, blurry hands and faces. & 12\% \\
\bottomrule
\end{tabular}
\end{table}

\begin{table}[h!]
\centering
\caption{ The comparison between the capabilities of different LVLM models}
\begin{tabular}{cccccc}
\toprule
\textbf{Model as Decoder } & \textbf{Image} & \textbf{Video} & \textbf{Pre-alignment} & \textbf{Combined- Fine-tuning} & \textbf{Reference} \\
\hline
Mini GPT-4 & \ding{51} & \ding{55} & - & \ding{55} & [29] \\
LLaVA & \ding{51} & \ding{55} & - & \ding{55} & [35, 37, 53] \\
Video-ChatGPT & \ding{55} & \ding{51} & - & \ding{55} & [31] \\
Video-Chat & \ding{51} & \ding{51} & \ding{55} & \ding{51} & [54] \\
Video-LLaMA & \ding{51} & \ding{51} & \ding{55} & \ding{51} & [32] \\
ImageBind-LLM  & \ding{51} & \ding{51} & \ding{51} & \ding{55} & [56] \\
Chat-UniVi  & \ding{51} & \ding{51} & - & \ding{51} & [36] \\
\color{green} AdCare-VLM (Ours)  & \color{green} \ding{51} & \color{green} \ding{51} & \color{green} \ding{51} & \color{green} \ding{51} & \color{green} - \\
\hline
\end{tabular}

\label{tab:example}
\end{table}

\begin{table}[h!]
\centering
\caption{The hyper-parameters for the AdCare-VLM}
\begin{tabular}{ccc}
\toprule
\textbf{Parameter Name} & \textbf{Pre-training} & \textbf{Fine-tuning} \\
\hline
Video Encoder & \multicolumn{2}{c}{LanguageBind-Video} \\
Image Encoder & \multicolumn{2}{c}{Languagebind-Image} \\
Text Encoder & \multicolumn{2}{c}{CLIP} \\
Optimizer & \multicolumn{2}{c}{AdamW} \\
Deepspeed & \multicolumn{2}{c}{zero, zero2, zero3} \\
Vision Select Layer & \multicolumn{2}{c}{2} \\
Epoch & \multicolumn{2}{c}{5} \\
Weight Decay & \multicolumn{2}{c}{0.01} \\
Warm up Ratio & \multicolumn{2}{c}{0.03} \\
Schedule & \multicolumn{2}{c}{Cosine Decay} \\
Tensor Size & \multicolumn{2}{c}{3 $\times$ 8 $\times$ 224 $\times$ 224} \\
Learning Rate & 1e-5 & 2e-5 \\
Batch Size & 64 & 128 \\
\hline
\end{tabular}

\label{tab:example}
\end{table}

\begin{table}[h!]
\centering
\caption{The fine-tuning results of AdCare-VLM with Chat-UniVi [36] on our LLM-TB-VQA dataset (7B model)}
\label{tab:univi-acc-score}
\setlength{\tabcolsep}{10pt}
\renewcommand{\arraystretch}{1.15}
\begin{tabular}{@{}l l c c@{}}
\specialrule{1.1pt}{0pt}{0pt}
\textbf{Model} & \textbf{Tuning Type} & \textbf{Accuracy (\%)} & \textbf{Score} \\
\specialrule{1.1pt}{0pt}{0pt}
 & Pre-trained & 50.3 & 3.27 \\
        Chat-UniVi (7B)                         & Regular     & 54.5 & 3.33 \\
                                 & LoRA        & 57.9 & 3.54 \\
\specialrule{1.1pt}{0pt}{0pt}
 & \color{green} Pre-trained & \color{green} 50.1 & \color{green} 3.1 \\
        \color{green} AdCare-VLM (Ours)                         & \color{green} Regular     & \color{green} 58.7 & \color{green} 3.4 \\
                                 & \color{green} LoRA        & \color{green} 61.2 & \color{green} 3.7 \\
\specialrule{1.1pt}{0pt}{0pt}
\end{tabular}
\end{table}

\begin{figure}[h!]
  \centering
  \includegraphics[width=0.8\textwidth]{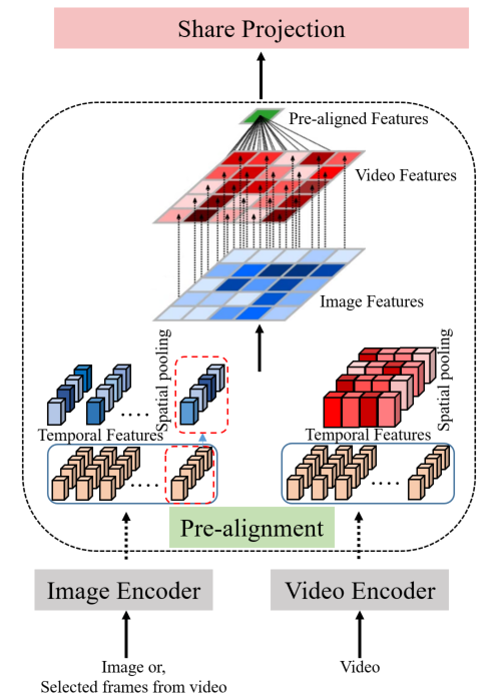} 
  \caption{The architectural overview of image and video pre-alignment prior to projection is provided to aid in the creation of a unified 
representation.}
  \label{fig:sample}
\end{figure}

\begin{figure}[h!]
  \centering
  \includegraphics[width=1\textwidth]{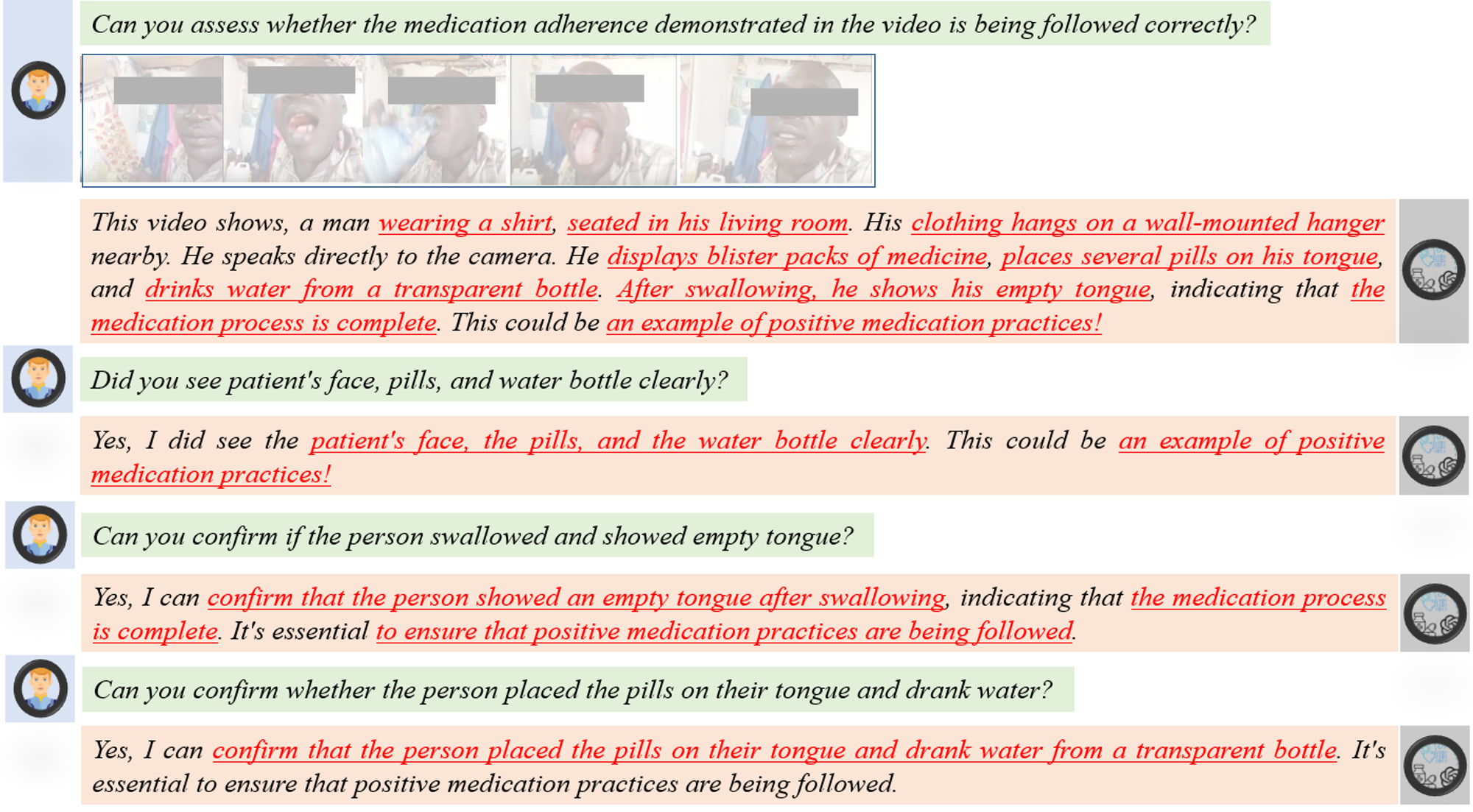} 
  \caption{AdCare-VLM-7B Model with LoRA fine-tuning on qualitative outcomes (A selection of key frames is presented (blurred to protect 
subject’s privacy), emphasizing critical features for {\color{green}positive} medication adherence, as indicated by the model, with red-underlined text). 
}
  \label{fig:sample}
\end{figure}

\begin{figure}[h!]
  \centering
  \includegraphics[width=1\textwidth]{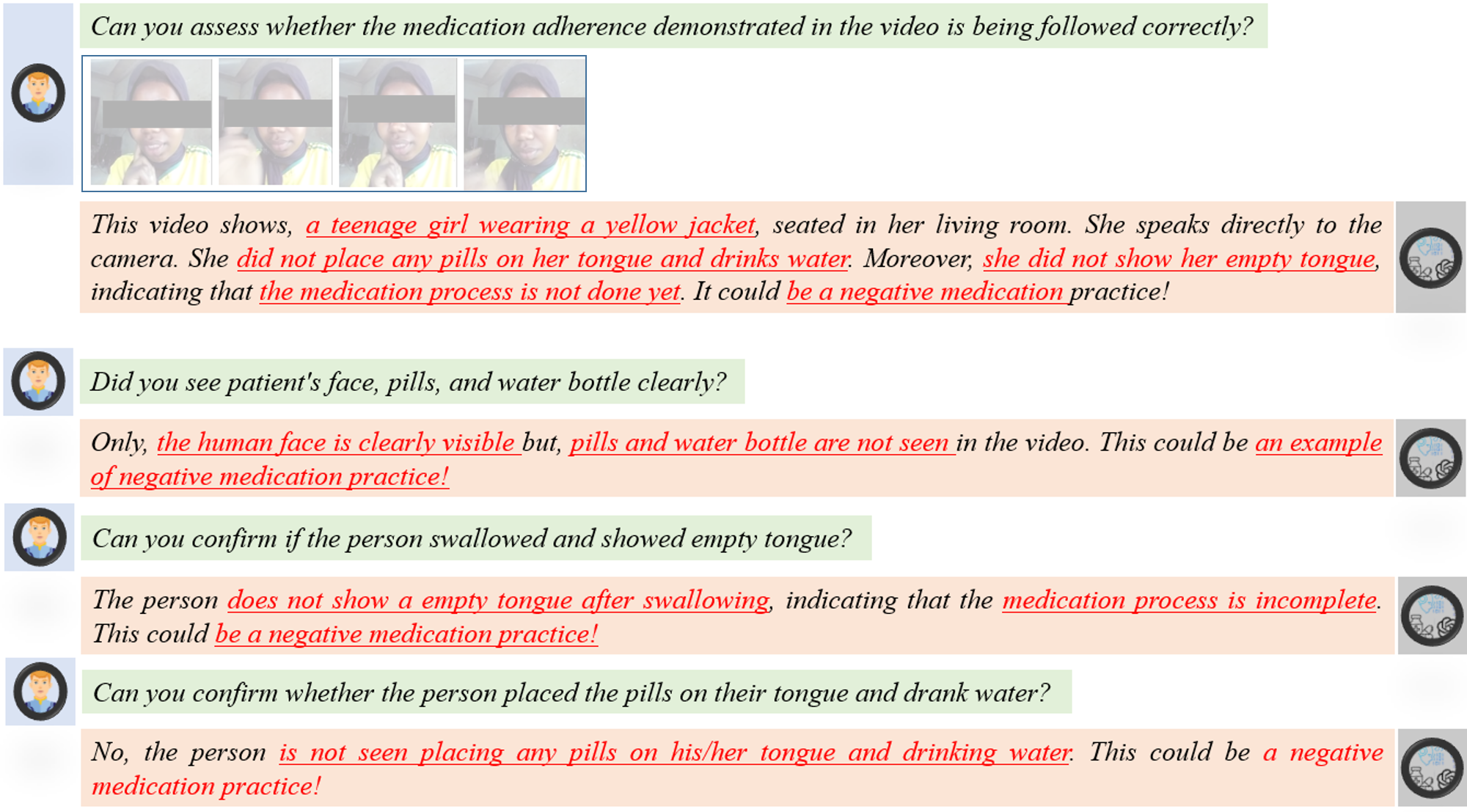} 
  \caption{AdCare-VLM-7B Model with LoRA fine-tuning on qualitative outcomes (A selection of key frames is presented (blurred to protect 
subject’s privacy), emphasizing critical features for {\color{green}negative} medication adherence, as indicated by the model, with red-underlined text). 
}
  \label{fig:sample}
\end{figure}

\begin{figure}[h!]
  \centering
  \includegraphics[width=1\textwidth]{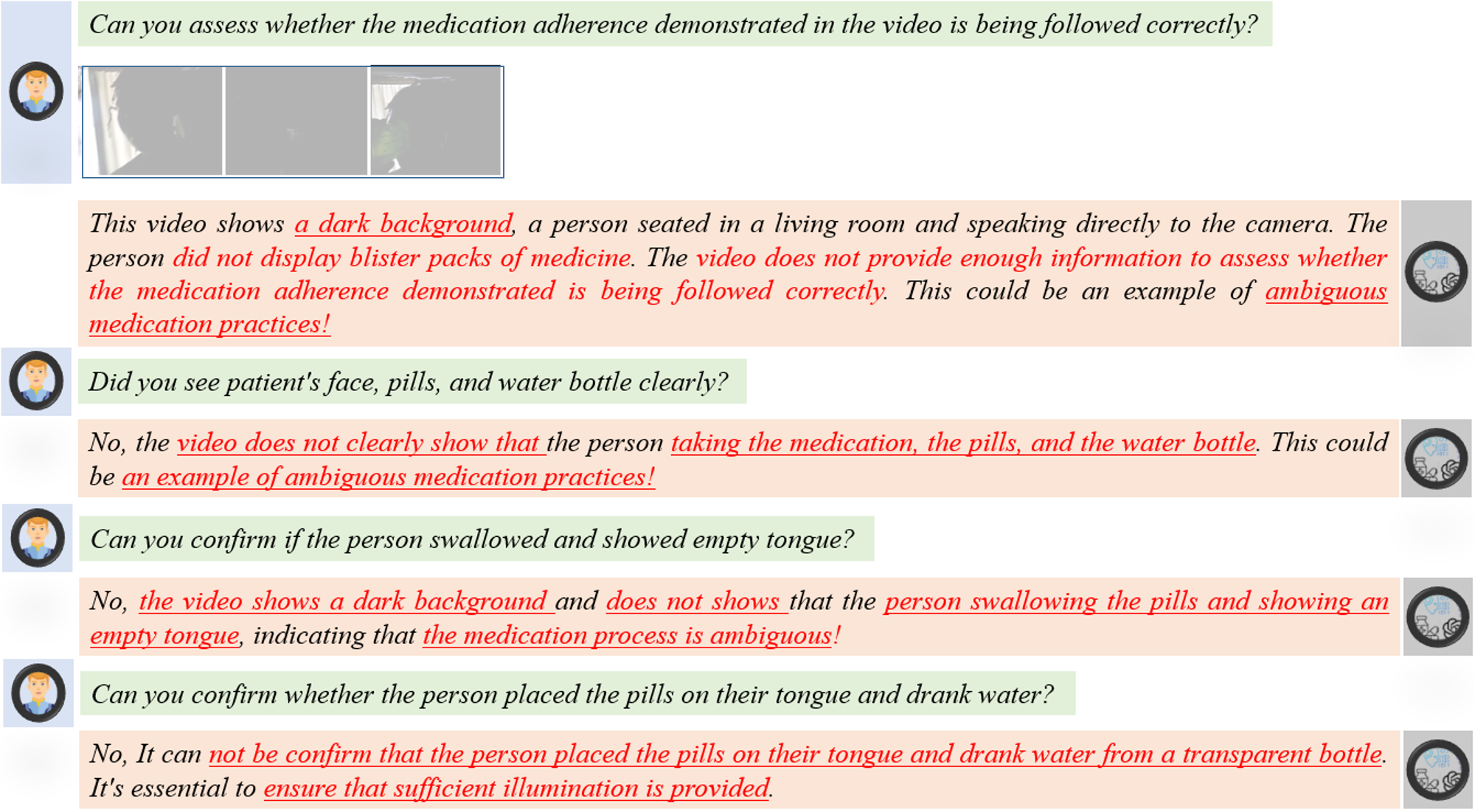} 
  \caption{AdCare-VLM-7B Model with LoRA fine-tuning on qualitative outcomes (A selection of key frames is presented (blurred to protect 
subject’s privacy), emphasizing critical features for {\color{green}ambiguous} medication adherence, as indicated by the model, with red-underlined text). 
}
  \label{fig:sample}
\end{figure}

\end{document}